# CHIP: Contrastive Hierarchical Image Pretraining

**Arpit Mittal**　　**Harshil Jhaveri**　　**Swapnil Mallick**　　**Abhishek Ajmera**

arpitm@usc.edu　hjhaveri@usc.edu　smallick@usc.edu　amajmera@usc.edu

## Abstract

Few-shot object classification is the task of classifying objects in an image with limited num- ber of examples as supervision. We propose a one-shot/few-shot classification model that can classify an object of any unseen class into a relatively general category in an hierarchically based classification. Our model uses a three-level hierarchical contrastive loss based ResNet152 classifier for classifying an object based on its features extracted from Image embedding, not used during the training phase. For our experimentation, we have used a subset of the ImageNet (ILSVRC-12) dataset that contains only the animal classes for training our model and created our own dataset of unseen classes for evaluating our trained model. Our model provides satisfactory results in classify- ing the unknown objects into a generic category which has been later discussed in greater detail.

## 1　Introduction

Recent research in the field of few shot object classification models [11], [8], [13] presume that one of the target classes is present in the query image. These models are not equipped to handle cases where the query image does not contain any of the target class objects along with incapability to categorize the classes. This limits the utility of these models making them incompetent for generic object classification tasks. We try to solve this problem by introducing a one-shot/few-shot learning model with Contrastive Learning [3], [9], [5] approach that can classify any unseen class into a relatively general category (Figure 1). For our project, we aim to classify only animal classes into more general categories. Our model exploits the class hierarchy using contrastive loss using CNNs with respect to the parent embeddings. Contrastive Learning is a method that is known to improve the performance of tasks by contrasting query samples against target labels. In doing so, this method helps to learn both the common attributes between data classes as well as the features that differentiate a data class from another. [1]

## 2　Related Work

Recently, a considerable amount of research has been conducted to add the aspect of generaliza- tion in few-shot learning models. However, most of these proposed networks, in spite of coming up with different novel approaches, fail to achieve good performance in generic few-shot object classification.

In order to improve generalizability, Jiawei Yang et al. [12] have used contrastive learning to annotate unlabelled data followed by latent augmentation methods consisting of k-means selection for training instances. Their key interest is to transfer semantic diversity from the training set to the augmented set. The features in the augmented sets are extrapolated and computed as a dot product and the most useful features are selected based on the magnitude of this dot product. However, the main problem with this approach is that it may be impactful only towards datasets with fewer classes but may not show significant improvements in generalization in datasets with relatively more classes. Alayrac et al. [1] achieve state-of-the-art results with regards to huge corpora of interleaved visual and text data with disparity in the method that context is sent for each example pair. A visual encoder is used to produce fixed length tokens which is then fed as additional input to weighted attention nets for embeddings generated from the text vectors. A frozen pretrained ResNet-50 architecture is used to preprocess image and video data. Since this architecture heavily utilizes transfer learning from large language models, it inherits their inductive biases. Moreover since the model aims at performing a series of tasks including classification, it lags in its performance as compared to other contrastive models.

---

[1]View the source code on GitHub: https://github.com/harshiljhaveri/CHIP/tree/main

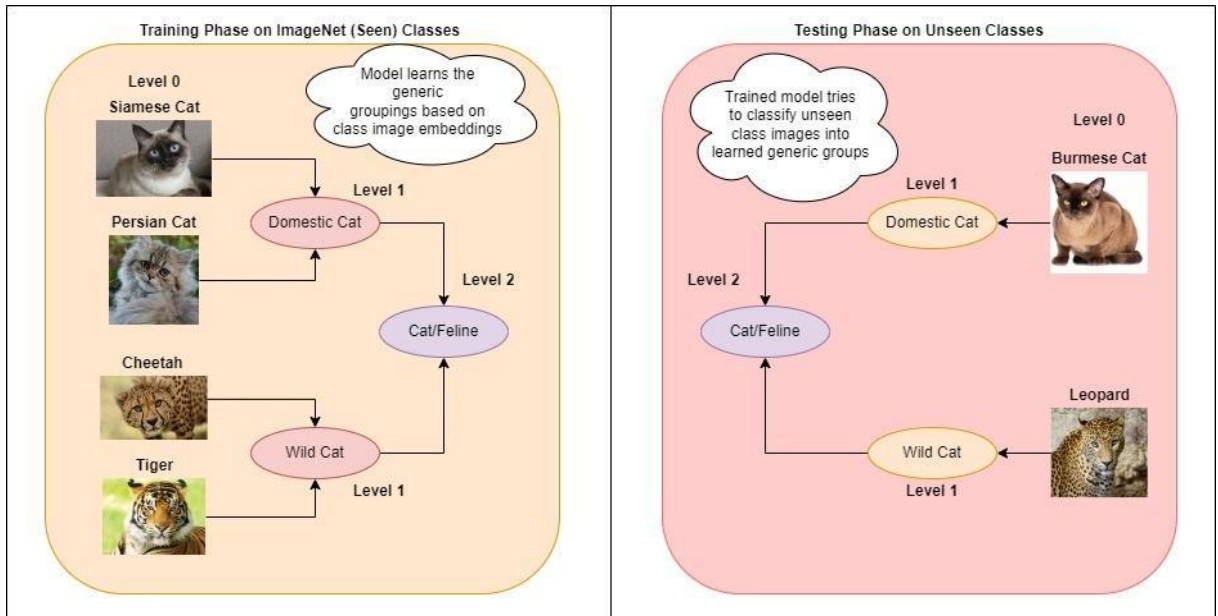
Figure 1: Pictorial representation of the aim of our project

Another attempt [7] to achieve generalization aims at addressing the issues of overlooked classes during classification of multi-label images into a single entity. At first, this method decomposes higher level images into smaller patches and annotates each of these using a fine grained vision transformer. Then a weighted tokening method is carried out to assign labels with minimum losses to each patch. Consequently, similarities are computed between query image patches and support images to assign closest classes. Tokens are assigned to each class based on temperature scaling while minimizing the cross entropy loss. Masks are applied to each patch in order to generalize the constraint and help it fit within the limited support image class options. However, the main drawback of this method is, the transformer is unable to de- fine strict decision boundaries where high feature similarity exists between classes and training data is limited in size.

We have tried to come up with a novel method to solve the problem of generalization in one-shot/few-shot learning that has been discussed in Section 4 in greater detail.

## 3 Dataset

The ImageNet [10] dataset consists of 14,197,122 images annotated according to the WordNet hierarchy and divided into 1000 classes. The ImageNet dataset is a part of the ImageNet Large Scale Visual Recognition Challenge (ILSVRC-12) which is a benchmark in object category classification and detection of images. Since the ImageNet dataset contains classes other than that of animals, we have segregated the images of all the animal classes from the dataset for our experiment. In doing so, we have got 366 animal classes with 1300 images in each class. We have used images from these classes in our training phase.

For testing our model, we have created our own dataset which contains 20 unseen animal classes that are not present in the ImageNet dataset. This dataset contains the following classes - *American Bobtail cat, American Paint horse, British Short- hair cat, Burmese cat, Camarillo White horse, Cat- fish, Crow, Cuckoo, Deer, Friesian horse, Giraffe, Ibis, Kingfisher, Kiwi, Parrot, Puffbird, Ragdoll cat, Rhinoceros, Sparrow and Tuna*.

## 4 Proposed Architecture

Our proposed architecture can be broadly divided into three phases:

### 4.1 Create Target Parent Image Embeddings (Phase 1):

The first phase of our architecture involves generating mean image embeddings of each ImageNet animal class using the pre-trained ResNet-152 [6] model and using unsupervised K-Means clustering to learn the class hierarchical structure (Figure 2).

**Algorithm 1:** Creating Target Parent Image Embeddings (*Phase* 1)
___
**Input**: Training Set $T$, Resnet-152 Image classifier $M$, Model pretrained weights $\Theta$
/* Load $M$ with $\Theta$ and remove last FC layer to get image embeddings from $M$, Embedding Dataframe $D$                                                                                        */
**Function** optimalKforCluster *(mean_image_embedding, minK, maxK)*:
    **for** $k$ *in range(minK, maxK)* **do**
        | KMeansMap [k] = Kmeans(mean_image_embedding, k)
    **end**
    $optimalK = silhouetteAnalysis$(KMeansMap)
    **return** $optimalK$
**end** $funtion$
**foreach** *Class $C$ in $T$* **do**
    **foreach** *Image $I$ in $C$* **do**
        | $imageEmbeddings = M$(Theta, I)
    **end**
    $mean\_image\_embedding\_for\_class = mean$(imageEmbeddings)
    $Level0ParentEmbeddingsperClass, D[C, 0]$ = mean_image_embedding_for_class
**end**
$Level1, 2ParentEmbeddingsperClass, D[C, 1] = KMeans(D[C,\text{level}],$
$optimalKforCluster$(D[C, level], minK, maxK))
**return** $D$

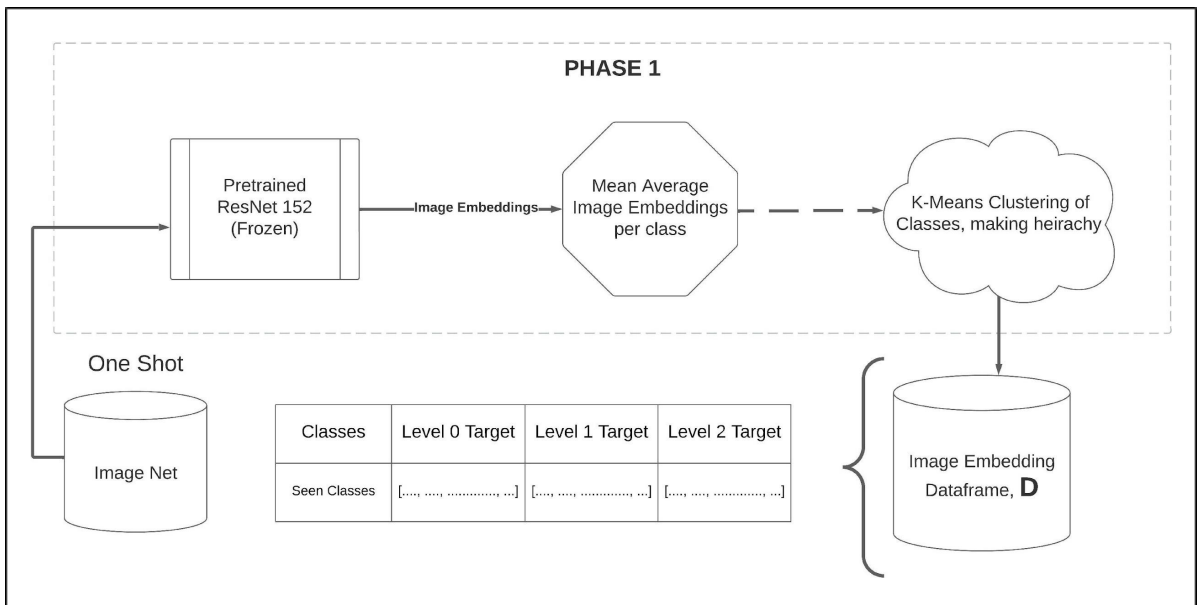

Figure 2: Phase 1 Network Architecture

**Algorithm 2:** One-Shot Hierarchical Model Learning ($Phase\ 2$)

**Input:** Training Data $Tr$, Validation Data $Va$, Test Data $Te$, Target Parent Embedding Dataframe $D$, Resnet-152 Image classifier $M$, Model pre-trained weights $\Theta$

$Tr$ = One random image from each class
$Va$ = 5 random image from each class
$Te$ = 5 random image from each class
$pretrainedResnet152Model = M(\Theta)$

**Function** contrastiveLoss *(imageEmbed, targetEmbed)*:
    $dist = CosineDistance$(imageEmbed, targetEmbed)
    $negDist = (\text{margin} - dist).\text{relu}()$
    $posDist = dist$
    $loss = Concat$(posDist, negDist).mean
    **return** $lossK$
**end** $funtion$

**Function** fineTuneModel *(model, trainData, valData, numEpochs)*:
    **for** *epoch in range(numEpochs)* **do**
        **for** *image, targetEmbed in trainData* **do**
            $embed = model$(image)
            $loss = contrastiveLossFn$(embed, targetEmbed)
            backPass
            updateGrads
        **end**
        **for** *image, targetEmbed in valData* **do**
            $embed = model$(image)
            $loss = contrastiveLossFn$(embed, targetEmbed)
        **end**
        $saveModelforEachEpoch()$
    **end**
    **return** $optimalPreTrainedModelForValData$
**end** $funtion$

**Function** testModel *(model, testData)*:
    **for** *image, targetEmbed in testData* **do**
        $imageEmbed = model$(image)
        $loss = contrastiveLossFn$(embed, targetEmbed)
        $cosineSim = CosineSimilarity$(imageEmbed, targetEmbed)
    **end**
    **return** $mean(loss), mean(cosineSim)$
**end** $funtion$

$Level\_HeirachialModel = fineTuneModel(\text{new } M(\Theta), (Tr, \text{level\_target\_embedding}), (Va, \text{level\_target\_embedding}), numEpoch, 0)$
**return** $Level\_HeirachialModel$

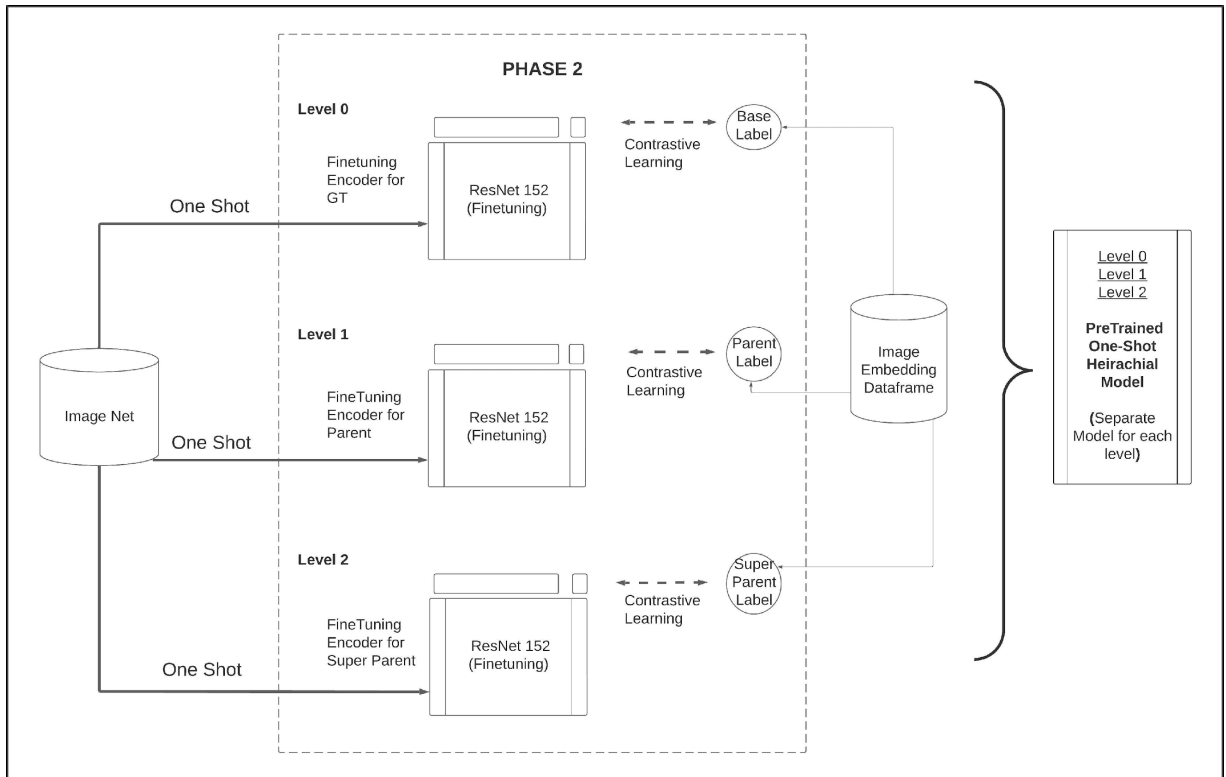

Figure 3: Phase 2 Network Architecture

The steps in Phase 1 are as follows:

1. At first, the mean image embeddings of each of the 366 ImageNet animal classes is computed by loading the ResNet-152 model with its pre-trained weights. So, we get 366 mean embeddings for 366 classes. They would be the target embeddings for the leaf node in our hierarchy and represent as Level 0 embeddings .
2. Using these mean embeddings from 1, these 366 animal classes are then grouped using K-Means clustering and the centroid embedding of each of the clusters is also calculated. Now we can associate each of the 366 classes to a cluster and its corresponding cluster centroid embedding. They would be the target embeddings for the parent node in our hierarchy and represent as Level 1 embeddings .
3. Next using the centroid embeddings from the previous step, these already formed clusters are further grouped into clusters using K-Means clustering and the centroid embeddings of these clusters are also computed. They would be the target embeddings for the super-parent node in our hierarchy and represent as Level 2 embeddings
In order to get the optimal number of clusters, Silhouette Analysis has been done both the times.
4. Finally, a mapping dataframe has been created that stores all the 366 classes, their corresponding mean class embeddings (Level 0) and maps each of these classes to their corresponding Level 1 and Level 2 clusters and their cluster centroid embeddings.

### 4.2 One-Shot Hierarchical Model Learning (Phase 2):

In the second phase of our architecture, a three- layered pre-trained ResNet-152 model is fine-tuned using one-shot learning approach (Figure3). The steps of Phase 2 are as follows:

1. In this phase, we have three separate models for three levels. We use ResNet-152 encoder for Level 0 Ground Truth. The Level 1 encoder is finetuned on parent clusters whereas the Level 2 encoder is finetuned on the super-parent clusters.
2. One image is taken from each of the 366 ImageNet animal classes and fed into all the three models (or layers).
3. For each image, an embedding is generated and Constrastive loss is calculated between the generated image embedding and the corresponding target embedding retrieved from the mapping dataframe generated in Phase 1.
4. Then we backpropagate and learn the respective

**Algorithm 3:** Few Shot Learning on Unseen Data (*Phase 3*)

**Input** : Unseen Data $UD$, Pre-trained Novel Models for Hierarchical Embeddings - $Level\_0\_HeirachialModel, Level\_1\_HeirachialModel, Level\_2\_HeirachialModel$

**Function** assignTargetLabelToNewClass (*classEmbeddings, level*):
  $distanceMatrix = CosineDistances$(classEmbeddings, D[level])
  $targetLabels$ = embedding with min distance for each class embeddings
  **return** $targetLabels$
**end** *funtion*

**Function** targetLabelsForUnseenClasses (*unseenClassData*):
  $classEmbeddings = Level\_HeirachialModel$(unseenClassData)
  $targetLables = assignTargetLableToNewClass$(classEmbeddings, level)
  **return** $targetLables$
**end** *funtion*

$D[UD, level] = targetLabelsForUnseenClasses(UD)$
$Level\_UnseenModel = fineTuneModel(Level\_HeirachialModel, D[UD,level], numepochs, level)$
**return** $Level\_UnseenModel$

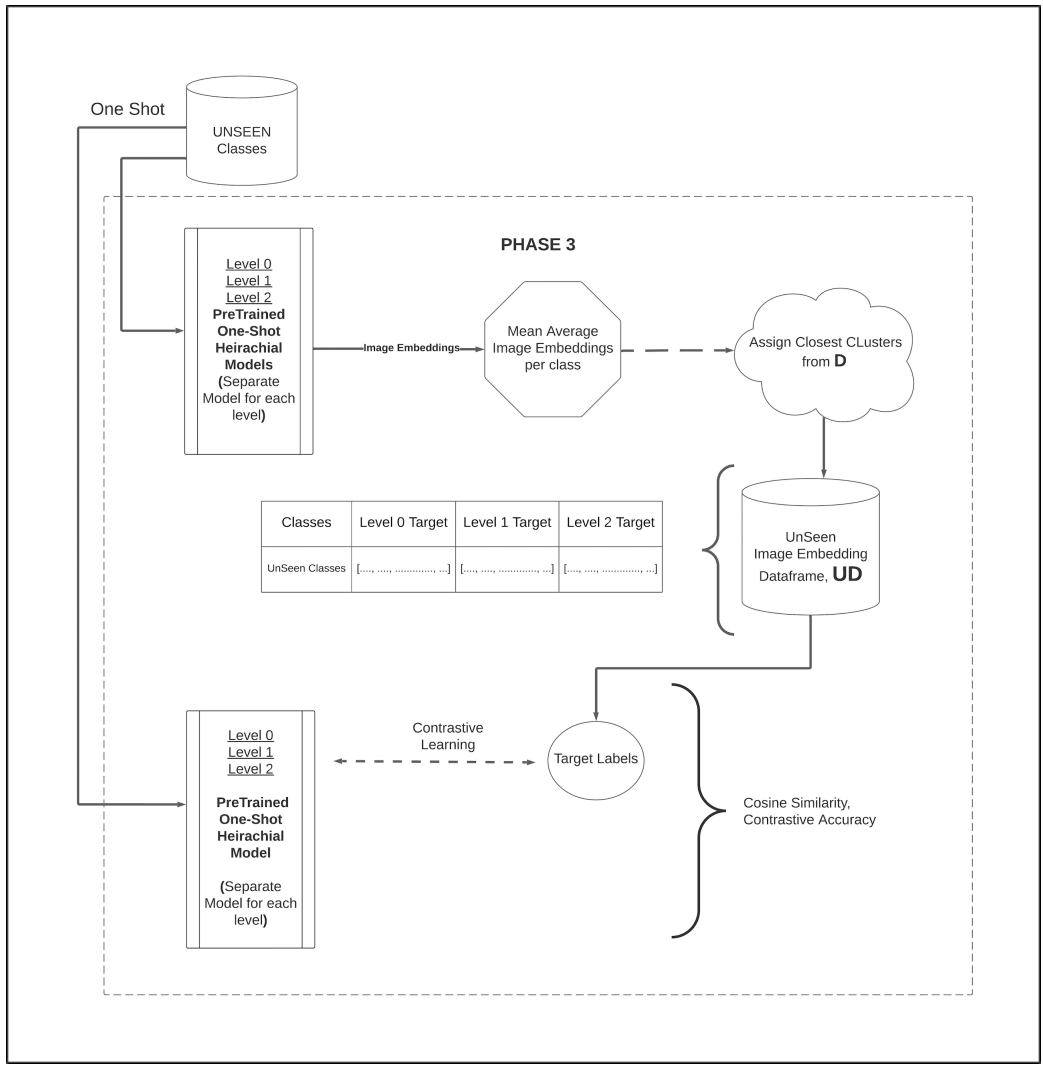

Figure 4: Phase 3 Network Architecture

weights.
5. Steps 2-4 are repeated for a set number of epochs and this is done for all the three separate models. Thus, we get pre-trained one-shot hierarchical model for each of the three levels.

### 4.3 Few Shot Learning on Unseen Data (Phase 3)

In this final phase, we use the pre-trained ResNet-152 model from Phase 2 and the clusters obtained from Phase 1 to classify our unseen data of 20 animal classes into similar clusters (Figure 4). This phase consists of the following steps:

1. First we feed the unseen class images to the three pre-trained ResNet-152 models from Phase 2 and obtain the mean class embedding for each level.
2. Then we assign target labels to each of the 20 unseen classes by calculating the cosine distance between the mean class embedding of the unseen classes and the mean embeddings obtained in Phase 1 dataframe. The class with the minimum cosine distance is assigned as the class of the unseen data. This done separately for each level.
3. Then we store the mapping of the unseen class embeddings in a new dataframe.
4. Following this, the three pre-trained one-shot models are finetuned using the mapping dataframe of the unseen classes over a set number of epochs.

## 5 Results

**Phase 1:** With 1300 images per class, it was computationally expensive to perform clustering and figuring out the optimal K for clustering. We performed 3 methods to get our clusters :
1. Mean Embedding of 1300 images per class.
2. Majority Pooling for 1300 data-points per class.
3. Mean Embedding of 10 images per class in succession, and then performing majority pooling of 130 embedding per class.

We perform a Silhouette analysis for the appropri- ate number of clusters and overall belongingness of each feature to a cluster. Other popular techniques like top-k accuracy and Cluster Purity were not used since we attempted to cluster features based on similarity rather than comparing with ground truth labels. With comparable Max Silhouette scores for all 3 choices, and method 1 being the most computationally feasible, we proceeded with Mean embedding of all images in a class.

**Level 1 results:** We implemented K-Means on a range of 50 to 200 clusters for Level 1 clustering and then performed silhouette analysis to evaluate the quality of the clustering results and selected the optimal k for level 1 of the hierarchy which came out to be 88 as in Figure 5.

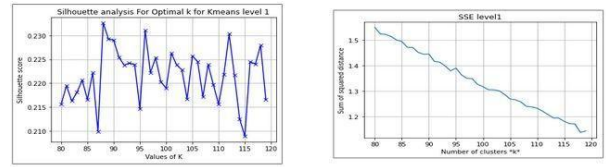

Figure 5: Level 1 results

**Level 2 results:** We decided to have minimum 7 clusters for level 2, and implemented K-Means on a range of 7 to 20 and based on the silhouette Analysis and Elbow Method, we decided to select 8 as the number of clusters for level 2.

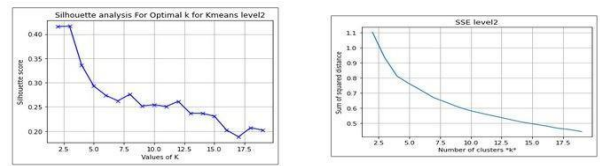

Figure 6: Level 2 results

For Phase 2 and Phase 3, we tried few shot, one shot and training on whole ImageNet dataset. One Shot training gave us much better performance, and we believe this is due to Variance in feature embeddings of the images when our ResNet-152 learns to classify images based on Feature Extraction with Contrastive Loss.

**Phase 2:** We used Cosine Similarity and Contrastive Accuracy to evaluate Phase 2 using validation and test data. Cosine similarity, in contrast to Euclidean distance, uses dot product rather than the magnitude of distances between the spatial features. As shown in Figure 7, we received an impressive accuracy of around 94% for each level and cosine similarity of over 0.85 for Level 0 and Level 1, and over 0.74 for Level 2. We believe, the drop in Level 2 cosine similarity is due to the fact that Level 2 is classifying more generalized features of the images at Super-Parent level (Table 1)

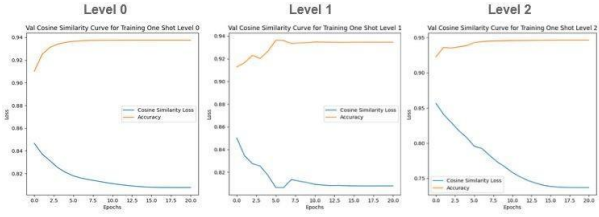
Figure 7: Phase 2 results

|  | Avg Loss | Avg Cosine Similarity |
|---|---|---|
| Level 0 | 0.035117637 | 0.92117435 |
| Level 1 | 0.03500075 | 0.5109769 |
| Level 2 | 0.03300126 | 0.5102735 |

Table 1: Phase 2 results for Seen Classes.

**Phase 3:** The evaluation metrics used for embeddings are Cosine Similarity for Pairwise Similarity between feature embedding of unseen images and the target embedding from Phase 2. While accuracy based on a threshold of pairwise similarity and precision-recall were calculated, they provided highly skewed results and hence were not considered further. Figure 8 shows the Training loss and Cosine Similarity for the 3 levels.

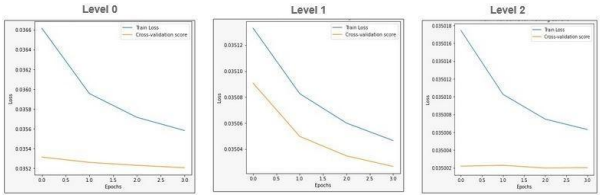
Figure 8: Phase 3 results

|  | Avg Accuracy | Avg Cosine Similarity |
|---|---|---|
| Level 0 | 93.8 | 81.5 |
| Level 1 | 93.9 | 82 |
| Level 2 | 95 | 74.5 |

Table 2: Phase 3 results for Unseen Classes.

There is a significant drop in cosine similarity for levels 1 and 2 when compared to level 0, which may be due to learning more generalized feature mapping. We also noticed a drop in Cosine Similarity for unseen classes, as compared to seen classes. We believe the reason for the drop in both cases is due to the nature of the unseen classes, i.e., the models were not pre trained on unseen classes to learn their features, as compared to pretrained model we used in Phase 2, which already had pretrained weights on ImageNet.

We tested our model on 20 unseen classes, and 14 of them were classified to their correct parent and super parent.

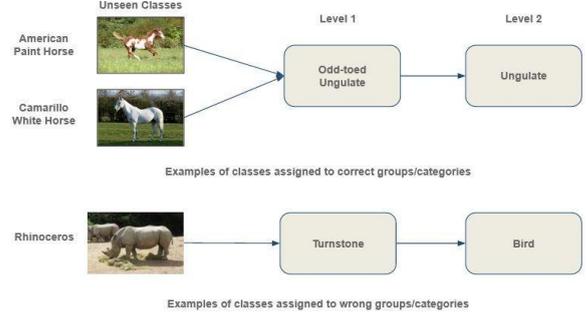
Figure 9: Phase 3 results

## 6 Conclusion and Future Work

In conclusion, the technique of using ground truth embeddings by clustering images based on feature similarity has been useful to the task of comparing feature extractions if image embeddings. Compared to the commonly performed classification tasks, comparing embedding features using contrastive learning required task adaptive ground truth labels. Thorough experimentation further proves that developing separate functions for each level of the hierarchy is better than using base levels to infer higher level labels. Further, not all evaluation metrics that commonly work for image classification tasks work in this domain since the points of comparison are pairwise similarities and cluster appropriateness rather than singular labels.

The proposed architecture has a lots of scope of improvement in each each of its phases. For Phase 1, we can try a more apt approach for clustering the Images based on Image Features, either by reducing its dimensions using PCA, using density based clustering like DBSCAN [4] or HDBSCAN [2] or other methods to cluster high dimension data. For Phase 2, we can use motivation from CLIP [9] model , and introduce Multi-Modal architecture , adding Textual, Prompt, and contextual features as well while training Level hierarchical models. For Phase 3, we can Better feature extractions and then modulate it with our Phase 2 models, similar to encoder-decoder architecture in transformers, to have the models learn features of our unseen classes in a broader way to improve similarity between our generalized Feature mapping of Hierarchical embeddings.